%% file: ms.tex
\documentclass[final]{cvpr}
\pdfoutput=1

\usepackage{times}
\usepackage{epsfig}
\usepackage{graphicx}
\usepackage{amsmath}
\usepackage{amssymb}
\graphicspath{{Images/}}
\usepackage{bm}
\usepackage{lipsum}
\usepackage{color}
\usepackage{booktabs}
\usepackage{algorithm}
\usepackage{multirow}
\usepackage{subfiles}

\usepackage[pagebackref=true,breaklinks=true,colorlinks,bookmarks=false]{hyperref}

\def\cvprPaperID{2955} % *** Enter the CVPR Paper ID here
\def\confYear{CVPR 2021}

\begin{document}

%%%%%%%%% TITLE
\title{It's All Around You: Range-Guided Cylindrical Network \newline for 3D Object Detection}

\author{Meytal Rapoport-Lavie\\
Tel-Aviv University\\
% Institution1 address\\
{\tt\small laviemeytal@mail.tau.ac.il}
% For a paper whose authors are all at the same institution,
% omit the following lines up until the closing ``}''.
% Additional authors and addresses can be added with ``\and'',
% just like the second author.
% To save space, use either the email address or home page, not both
\and
Dan Raviv\\
Tel-Aviv University\\
%First line of institution2 address\\
{\tt\small darav@tauex.tau.ac.il}
}

\maketitle

\input{00_abstract.tex}
%%%%%%%%% BODY TEXT

\input{01_introductionB.tex}

\input{02_related_works}

\input{03_challenges}

\input{03_method.tex}

\input{04_experiments.tex}

\input{05_conclusions.tex}

{\small
\bibliographystyle{ieee_fullname}
\bibliography{06_references}
}

\end{document}

%% file: 00_abstract.tex
\begin{abstract}
Modern perception systems in the field of autonomous driving rely on 3D data analysis. LiDAR sensors are frequently used to acquire such data due to their increased resilience to different lighting conditions. Although rotating LiDAR scanners produce ring-shaped patterns in space, most networks analyze their data using an orthogonal voxel sampling strategy.
This work presents a novel approach for analyzing 3D data produced by 360-degree depth scanners, utilizing a more suitable coordinate system, which is aligned with the scanning pattern. Furthermore, we introduce a novel notion of range-guided convolutions, adapting the receptive field by distance from the ego vehicle and the object's scale.
Our network demonstrates powerful results on the nuScenes challenge, comparable to current state-of-the-art architectures. The backbone architecture introduced in this work can be easily integrated onto other pipelines as well. %\footnotemark. 

%\footnotetext{Our code will be publicly available upon publication.}
%3D data is a fundamental and critical ingredient in modern perception systems. The geometric structure is robust to lighting conditions and improves object-background separation, which explains its outstanding performance for perception in automobile autonomous driving systems. 
%LiDAR scanner generates rings-shape patterns in space, yet we are used to resampling and analyzing the data in a voxel-like orthogonal sampling strategy. This work provides a novel approach for analyzing 3D data produced by 360-degree depth scanners. We show that we can use more suitable coordinate systems aligned with the scanning patterns and present a novel concept of range-guided convolutions well-suited to the adaptive voxel sizes in space.
%We show powerful results, comparable to state-of-the-art architectures, for our perception system on the nuScenes challenge. The backbone of this work can be easily added to other pipelines as well. 
\end{abstract}
 

%% file: 01_introductionB.tex
\section{Introduction}\label{sec:intro}
% The field of autonomous driving is gaining more and more attention over the last years. One main 

% Visual perception systems for the autonomous driving effort must be robust. It ought to be able to operate in any lighting scenario, day or night, in harsh weather, for any requisite duty. The addition of other sensors beyond the standard RGB cameras, such as LiDAR, is one important step towards achieving this objective. 

Robustness is a crucial requirement for visual perception solutions in autonomous driving systems. Such systems must be resilient to various lighting scenarios and withstand harsh weather conditions without compromising their performance. Therefore, extending the traditional RGB cameras with additional sensors, such as LiDAR, is a vital step towards achieving this goal.
%--------------------------------------------------------------------
\input{Images/teaser/teaser_fig}

%--------------------------------------------------------------------
The inclusion of depth information, which allows capturing the three-dimensional structure of the vehicle's environment, is a key feature for ensuring robustness while maintaining high accuracy. Modern LiDAR acquisition sensors provide meaningful information not just for avoiding imminent collisions but to perceive the environment as good as image-based data and even surpass it under poor lighting conditions.

%Utilizing low-resolution scanners may be sufficient for avoiding imminent collisions. Nevertheless, the information collected by modern LiDAR acquisition sensors can be at least as meaningful in depicting its environment as image-based data. Whereas under poor lighting conditions, it surpasses it substantially.

% Range information, which helps to capture the structure of the environment, holds an important key for achieving robustness while retaining the required high accuracy \cite{rangercnn}. When the resolution of the scanner is low, it can help to warn of approaching obstacles, but modern LIDAR acquisition sensors have sufficient resolution to capture meaningful information about the scene as good as done with images, in many cases better – especially in hard lighting conditions. Nevertheless the 3D data is irregular and sparse. This raises questions about how best to represent it, especially when dealing with large 3D scenes.

% In autonomous driving, we have seen a variety of techniques copying with a cloud of points, such as using PointNet \cite{pointnet} or pointNet-like architectures \cite{frustrum, pointrcnn}, but the most common method amongst state-of-the-art works, both for accuracy and efficiency, is to quantize the space according to the Cartesian coordinate system \cite{avod,mv3d, partA2, voxelnet, pixor}.

In the field of autonomous driving, there is a variety of methods to handle point clouds. The most common and efficient approach is to transform the point-cloud into a regular representation, such as 2D bird-eye-view (BEV) voxels \cite{avod,mv3d, pixor, pointpillars} or 3D voxels \cite{ partA2, voxelnet, pointrcnn,second}.  Following the architecture of PointNet and PointNet++ \cite{pointnet, pointnet++}, some works have proposed replacing the voxel-based method with a point-based one. This approach, however, was found to be computationally intensive. Recent papers have suggested combining both point-based and voxel-based methods into a single network \cite{pvrcnn, rangercnn}. However, whether combining point-based methods or not, in 2D or 3D, these methods use the Cartesian coordinate system as their grid.

% In autonomous driving, we have seen a variety of techniques copying with a cloud of points. The most common way and also the most efficient one is to transform the point cloud into a regular representation such as 2D bird-eye-view (BEV) voxels \cite{avod,mv3d, pixor, pointpillars} or 3D voxels \cite{ partA2, voxelnet, chen2019fast,second}. Following the architecture of PointNet and PointNet++ \cite{pointnet, pointnet++} some works tried to replace the voxel-based method with a point-based one \cite{frustrum,f-conv, pointrcnn,std}. Recently we witness a number of methods that are trying to combine them both, the point-based and voxel-based into a single network \cite{pvrcnn, rangercnn}. However, whether a voxel-based method or a combined method is used, either in 2D or 3D, it is done by the Cartesian coordinate system.

In this paper, we argue that projecting the points onto a Cartesian space disregards the circular nature of the raw data provided by a rotating emitter. We show that in order to perceive and preserve the additional information gained from a LiDAR sensor, one may leverage a more appropriate system, namely the Cylindrical coordinates system.
Furthermore, the Cylindrical coordinates system allows us to align the entire network towards the sensor's point of view. Intuitively, the closer an object is to the ego vehicle, the more points correspond to it in the point cloud. Therefore we suggest adjusting the size of the voxels to the distance from the sensor, as is done naturally by the Cylindrical space.
To do so, we propose a novel guiding unit, which orchestrates convolutions by their range from the ego vehicle. We further explore the challenges for a "self-oriented" network to learn angle and velocity information and present the necessary solutions to overcome them successfully.

\vspace{0.5cm}

Our main contributions are as follows:
\begin{itemize}
  \item We present the first end-to-end cylindrical coordinates perception architecture for autonomous systems, including orientation and velocity estimation.
  \item We present a novel range-guided convolution block to adjust the network's receptive field according to an object's range and scale, and adapting its learned features by its distance from the ego-vehicle.
  \item We provide new insights, challenges and solutions for training deep networks in cylindrical coordinates.
  \item Our network achieves comparable results to current state-of-the-art networks on the competitive nuScenes 3D object detection benchmark.
\end{itemize}

% Autonomous driving is a growing field/ 3D object detection is a must in order to map the surrounding of the vehicle/ Cameras-only methods are not safe enough/ LiDAR is needed/ Point cloud is irregular and sparse/ The common representation of point cloud, especially in big outdoor scenes, is voxelization/ Divide the map into voxels and then average all point's features/ The voxelization is done in Cartesian coordinates, similar to pixels in 2d/ But the LiDAR data is characterized in rings in the field, with strong correlation to range from ego vehicle/ Especially when one combines close sweeps in time (like nuScenes that has each 0.05s) Better resolution for close points/ If you Average on a part of an arch in Cartesian, you'll get an average point that doesn't even exist, in oppose of if you did the average in Spherical coordinates/ 

%% file: Images/teaser/teaser_fig.tex
\begin{figure}[htb!]
\begin{center}
% \textbf{Mean Points per Voxel}
\includegraphics[width=1.0\linewidth]{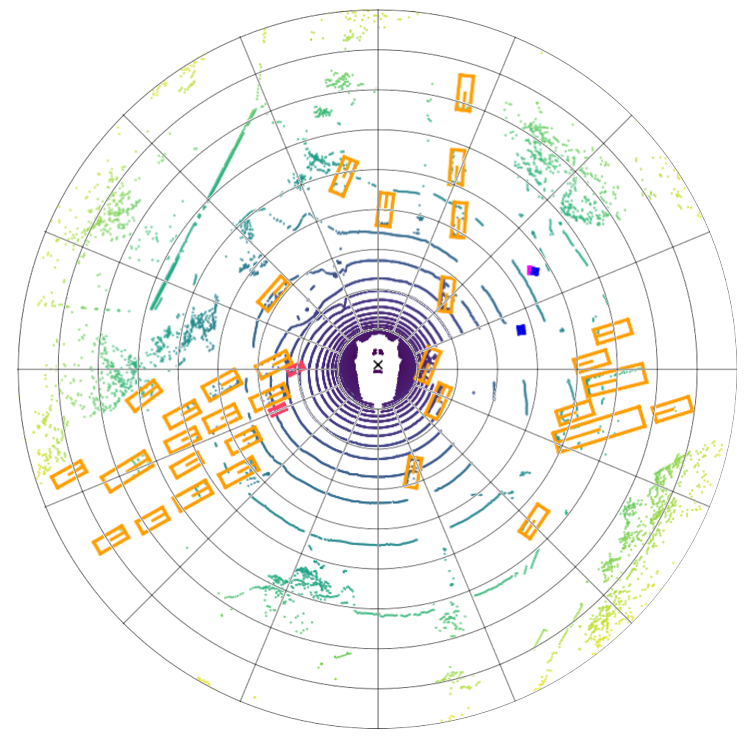}
\caption{Top-View of a nuScenes dataset sample with Ground Truth detections. First, it can be easily observed that the LiDAR sensor has a ring-shaped pattern output. Second, the Cylindrical Coordinate system maintains 9 neighbors and 27 neighbors per voxel (including itself), in 2D and 3D respectively, the same as the classical Cartesian system. This enables us to employ the classical convolution layers without almost any alteration.}
\end{center}
\label{fig:teaser}
\end{figure}

%% file: 02_related_works.tex
\section{Related works}

\paragraph{Cartesian Based 3D Object Detection Methods:}
Most state-of-the-art methods for 3D object detection project the point clouds onto a Cartesian coordinate grid. Earlier approaches employed the BEV projection to utilize efficient 2D convolution layers to process the point cloud data \cite{avod,mv3d}. In a later work, VoxelNet \cite{voxelnet} was first to introduce 3D voxels in the field of object detection, whereas SECOND \cite{second} improved this method by applying sparse 3D convolutions to accelerate running time. Most cutting edge methods \cite{partA2,pvrcnn}, including the previous and current nuScenes' state-of-the-art, CBGS and CenterPoint \cite{cbgs,centerpoint}, have adapted the practice of processing the point cloud through sparse 3D convolutions after projecting it to the traditional Cartesian voxel space.

% Earlier approaches used the bird-eye-view (BEV) projection in order to view the points in 2D, for efficient processing with common 2D convolution layers \cite{avod,mv3d}. VoxelNet \cite{voxelnet} was the first to use 3D voxels for object detection, whereas SECOND \cite{second} extended this by employing sparse 3D convolutions which accelerated performance time. Many cutting edge methods followed \cite{second} and used the sparse 3D convolution \cite{partA2,pvrcnn}, including nuScenes' state of the art CBGS and CenterPoint \cite{cbgs,centerpoint} which also uses the traditional Cartesian coordinates for voxelization.

%----------------------------------------------------------------------------------------------
\input{Images/points/fig_point_voxel}
%----------------------------------------------------------------------------------------------

\paragraph{Cylindrical / Spherical Based Methods:}
Only recently, PolarNet~\cite{polarnet} has suggested quantizing LiDAR data in a non Cartesian coordinate system. Their Polar-BEV network, designed for semantic segmentation on the SemanticKITTI \cite{sementickitti} challenge, have shown improvement with respect to traditional Cartesian methods. Unlike semantic segmentation tasks, 3D detection challenges, such as nuScenes, require the processing of additional object orientation and velocity features. As direction information plays a significant role in such tasks, they are directly affected by the change in the coordinate system and therefore need to be treated with special care.

%Although LiDAR sensors produce ring-shape point clouds, it was only  recently that we encountered a work that suggested quantizing it in a different coordinate system. PolarNet \cite{polarnet} is a network designed for semantic segmentation on the SemanticKITTI \cite{sementickitti} dataset. The network suggests quantizing the point cloud into polar-BEV and demonstrates improvement over traditional Cartesian methods. In the segmentation task, one is asked to output point's class, which is hardly affected by the choice of coordinates. n the nuScenes benchmark, as we will describe in detail later, one is requested to output additional information regarding the detected objects, which is much affected by the chosen coordinate system and hence requires special care.

\input{Images/cylin/theta_center}
\input{Images/Arch/tex/flow_wider}
%----------------------------------------------------------------------------------------------
\paragraph{3D Object Detection Methods with Focus on Range:}
LaserNet~\cite{lasernet} has introduced the concept of range-image input to the field of 3D detection, naturally emphasizing an object's distance from the ego vehicle as one of its key features. Despite its computational efficiency, it was generally outperformed by voxel-based methods. A later publication~\cite{range_google} suggested a method to address range-image's problem of scale variance, the fact that near distance objects appear larger in the image view. They applied dilated convolution \cite{dilated1,dilated2} directly to the range-image input, adjusting each pixel's dilation rate as a function of its range.

% In a recent paper, RangeRCNN~\cite{rangercnn} argued that although range-image is insufficient by itself, incorporating it as a source of initial feature extraction, before transforming the data into BEV architectures can be effective. This method of enhancing BEV with range-image features achieves state-of-the-art results on the KITTI dataset~\cite{kitti}.

In two recent papers CVCNet and RangeRCNN~\cite{cvcnet, rangercnn} tried to benefit from both range view and BEV. CVCNet~\cite{cvcnet} proposed a pair of cross-view transformers to transform the feature maps into the other view and offered a new voxel representation, Hybrid-Cylindrical-Spherical (HCS) voxels, which enables them to extract features for both Range view and BEV in a unified coordination system. RangeRCNN~\cite{rangercnn} argued that although range-image is insufficient by itself, incorporating it as a source of \emph{initial} feature extraction, before transforming the data into BEV architectures can be effective. This method of enhancing BEV with range-image features achieves state-of-the-art results on the KITTI dataset~\cite{kitti}.

In our work, we use range information to guide the 3D convolution layers. In this manner, we derive additional benefit from its data as an initial feature, allowing the network to comprehend objects differently depending on their distance from the ego vehicle. Furthermore, we use range information to manage the receptive field of the voxels, for overcoming their scale variance in the Cylindrical coordinate system.

%% file: Images/points/fig_point_voxel.tex
\begin{figure}[htb!]
\begin{center}
\textbf{\space\space\space\space\space\space\space\space\space\space Mean Points per Voxel}
\includegraphics[width=7.8cm, height=6.1cm]{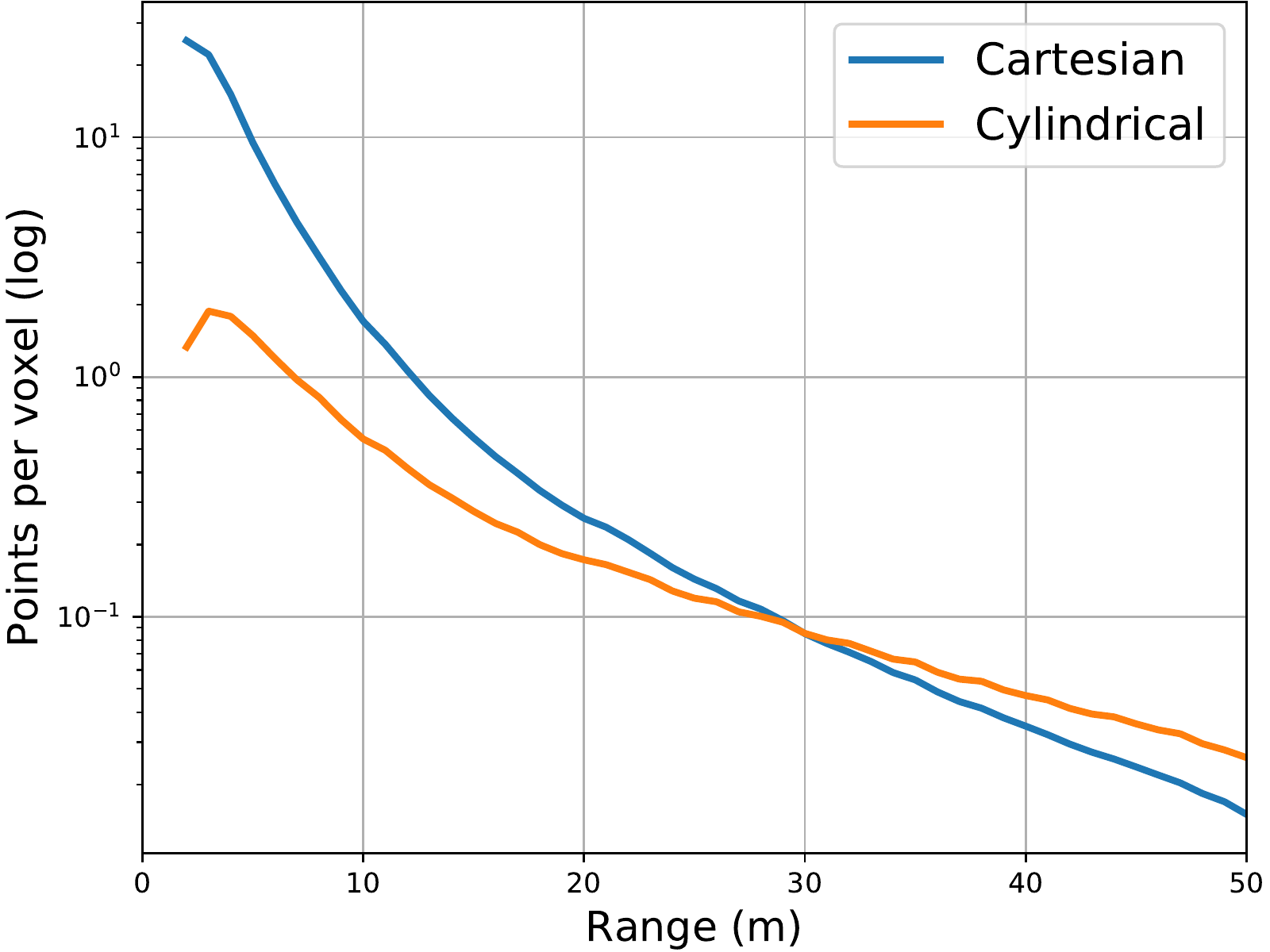}
    \caption{Mean number of points per voxel (logarithmic) as a function of range. There is a high inverse correlation between the number of points and their range from the sensor. Since the Cartesian voxels are of the same size in all ranges, we lose precision in the nearby range. On the other hand, in Cylindrical coordinate we receive a better spatial spread.}
\end{center}
\label{fig:avg_points}
\end{figure}

%% file: Images/cylin/theta_center.tex
\begin{figure*}[htb!]
\begin{center}
\textbf{Cylinder Map to Input Array for Objects with same \bm{$\theta_{dir}$}}
\includegraphics[width=0.9\linewidth]{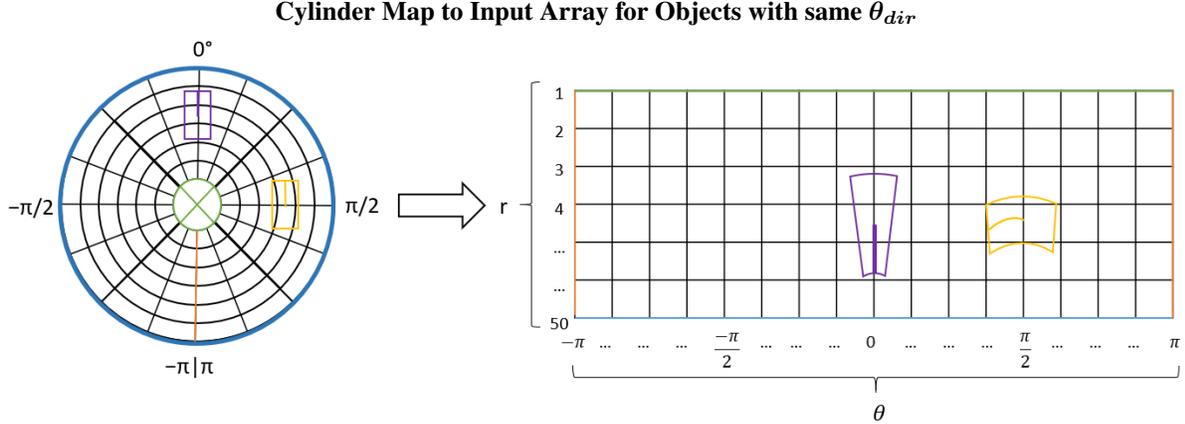}
\caption{Illustration of the Cylindrical map translated into the network's 2D array. Here shown two identical cars placed at the same distance from the ego vehicle but have different $\theta_{center}$. Since we are working from the sensor's point of view, we can see that it translates differently to the 2D array. The network observes the yellow car from the side as the purple car from behind, just as the sensor/driver does.}
\label{fig:dif_theta}
\end{center}
\end{figure*}

%% file: Images/Arch/tex/flow_wider.tex
\begin{figure*}[bt!]
\begin{center}
  \includegraphics[width=17cm, height=9.6cm]{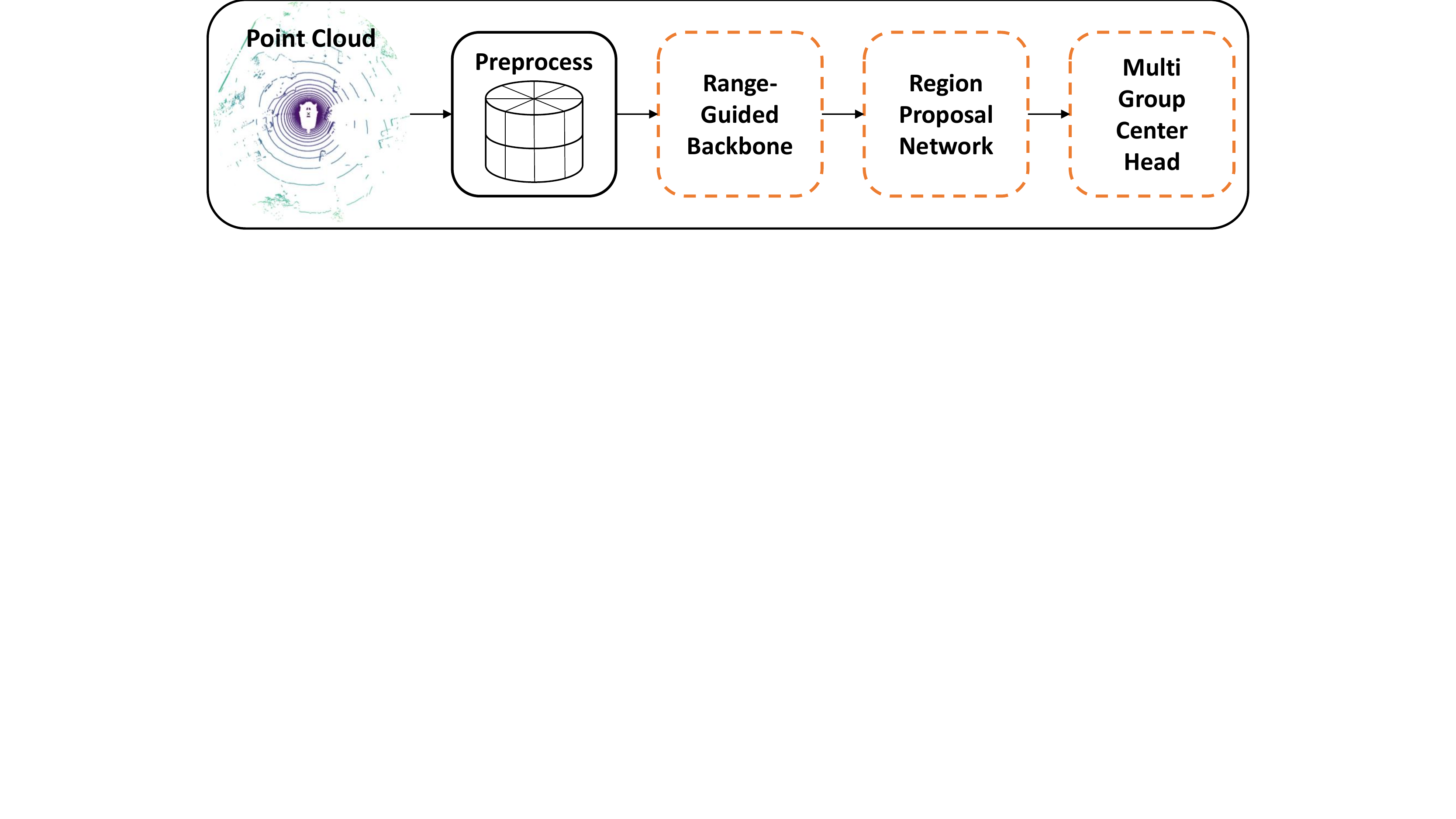}
\end{center}
 \vspace{-6.8cm}
  \caption{{\bf A diagram illustrating the main components of the network architecture.} Orange dashed lines indicating deep networks.}
\label{fig:flow}
\end{figure*}

%% file: 03_challenges.tex
\section{Challenges in Cylindrical Coordinates} \label{challenges}

The Cylindrical-coordinate grid differs from the Cartesian in many aspects. First and foremost it is a \textit{self-oriented} system. Consequently, identical objects placed at different positions relative to the sensor location will be presented differently along the network's channels. 
For this reason, in order to adapt the network to the sensor's point of view, certain challenges have to be addressed.

% The cylindrical coordinate system differs from the Cartesian system in many aspects. The first and most significant, from which many others are also derived, is that it is a \textit{self-centered} system.
\input{Images/Arch/tex/backbone_fig}
\input{Images/Arch/tex/weight_fig}
%-------------------------------------------------
In contrast to Cartesian coordinates BEV, the size of a Cylindrical voxel varies depending on its range. Thus two identical objects can spread over a different number of voxels, depending on their center location relative to the ego vehicle. Likewise, the orientation of an object toward the ego vehicle might also affect the number of voxels it occupies in the Cylindrical grid. In general, objects spread on a larger amount of voxels, require a wider receptive field in order to detect them. 

% In oppose to Cartesian coordinates bird-view, in Cylinder coordinates the size of a voxel is not uniform across all ranges. Meaning two identical objects can spread over a different number of voxels depending on their center location relative to the ego vehicle. Both the angle and range  affect the mapping of the object. Generally speaking, the closer an object is, the more voxels it will occupy thus requiring a larger reception field.

The Cartesian and Cylindrical-coordinate systems differ in the way they perceive the orientation of objects. The network's point of view can be equated with the perspective of a passenger in the ego-vehicle, observing two identical cars, one of which is in front of the vehicle while the other is on its side. As illustrated in Figure \ref{fig:dif_theta}, both yellow and purple cars, are of the same size, facing the same way, and at the same range. However, the difference in their $\theta$-centers causes them to be mapped differently to the network's input arrays with an emphasis on the direction. As shown, the purple car and the yellow ones are no longer facing the same direction in the 2D input array of our network. In order for the network to learn its orientation accordingly, it should be modified in a proper manner.

% As shown, in the array the purple car is facing outward whereas the yellow one is turned to the left. In order for the network to learn their orientation accordingly, it should be modified in a proper manner.

The same principle applies to the velocity's direction. For example, an object moving along a certain axis will seem to the Cartesian-based network to be moving along that axis regardless of its location on the map. This behavior manifests because the axes of Cartesian coordinates are aligned to those of every other object on the map. However, for the Cylindrical coordinates, this is not the case, and therefore additional adjustments are necessary.

%One can imagine how the network would view each object as if one was sitting within the ego vehicle and turning one's head in the direction of the object. This means that two identical cars, for example, with the same scales and moving direction, will look dissimilar in Cylindrical coordinates if they do not share identical center coordinates as we can see in Figure \ref{fig:dif_theta}. 

% The same principle applies to the velocity's direction. A bus traveling along a certain axis, for example, will seem to the Cartesian based network to be moving along that axis regardless of where it is located on the map. This occurs since the direction of the X and Y axes are aligned with those of all other objects on the map. However, for the cylindrical coordinates, which as we mentioned are a self-oriented system, this is not the case.  

Additionally, since the $\theta$ axis is inherently cyclical ($-\pi$ is also $\pi$), objects may spread simultaneously over voxels at both the beginning and the end of the network's input arrays. This is a further deviation from the Cartesian system that needs to be taken into account in order to accomplish the transition between the two systems.

In conclusion, transferring from a standard coordinate system towards a self-oriented one reveals our networks' unique challenges. To take full advantage of the many benefits offered by Cylindrical coordinates, fundamental changes are needed throughout the network's architecture and the outputs that are being learned.
%-------------------------------------------------------------------------
% Moreover, as a result of the circularity of the $\theta$ axis, an object can spread over both the start and end of the map, since in fact, $-\pi$ is also $\pi$. This principle does not apply to the Cartesian coordinate system and needs to be addressed.
%------------------------------------------------------------------------- 

%% file: Images/Arch/tex/backbone_fig.tex
\begin{figure*}[htb!]
\begin{center}
  \includegraphics[width=\linewidth]{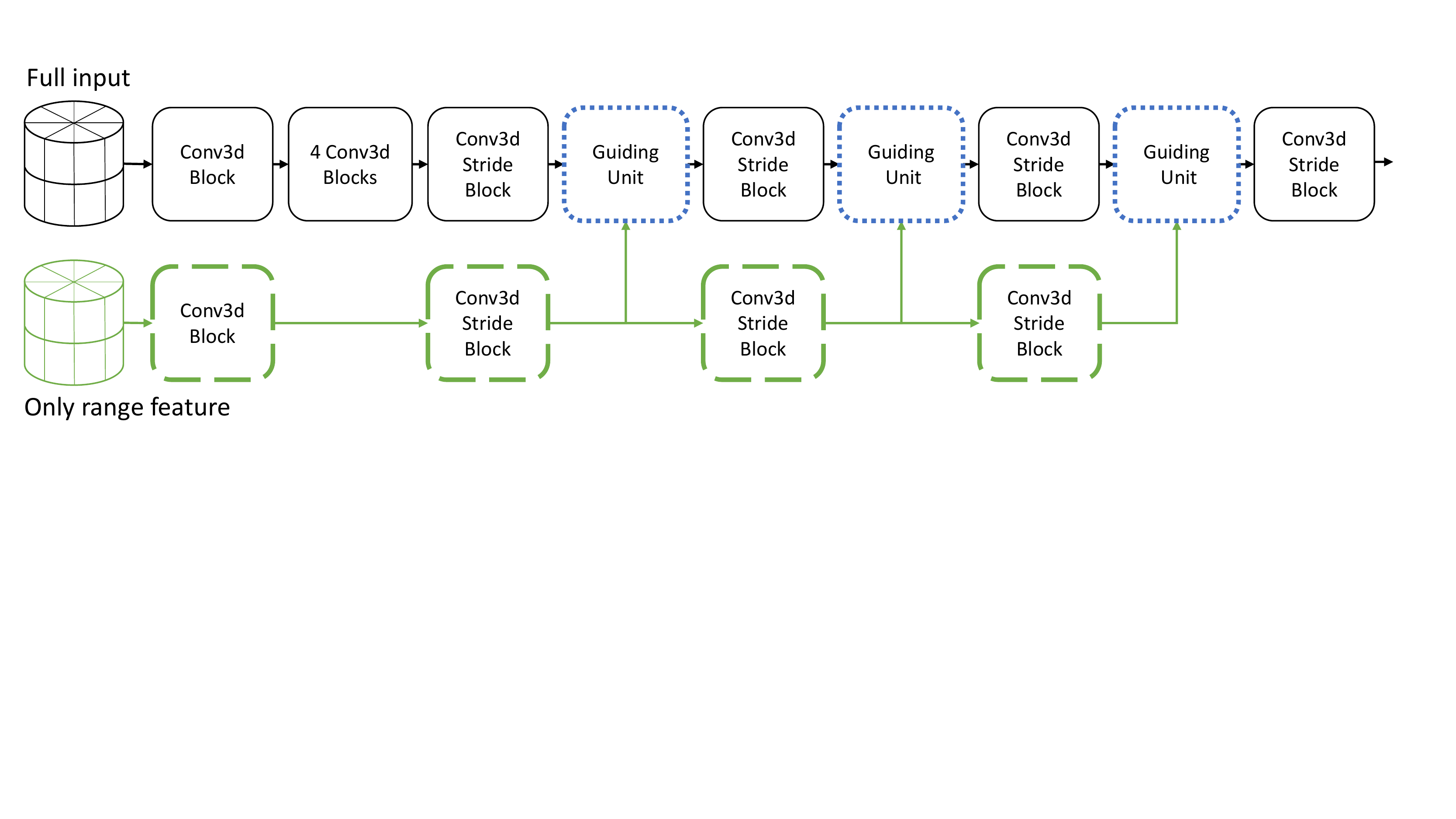}
\end{center}
\vspace{-4.8cm}
  \caption{{\bf Illustration of the Range-Guided Backbone:} The \emph{guided} backbone is shown in black, the range \emph{guiding} backbone is shown in a green dashed line and the guiding unit are shown in blue doted lines. We added a parallel range feature branch in order to guide a simple backbone consisting of convolution layers.}
\label{fig:only_backbone}

\end{figure*}

% The range \emph{guiding} backbone is made of a convolution pipeline sourced by range only matrix (meaning only the range feature of each voxel), controlling the number and weights of consecutive convolutions running on the main pipeline. The range \emph{guiding} backbone

%% file: Images/Arch/tex/weight_fig.tex
\begin{figure*}[t!]
\begin{center}
  \includegraphics[width=16.3cm, height=9.4cm]{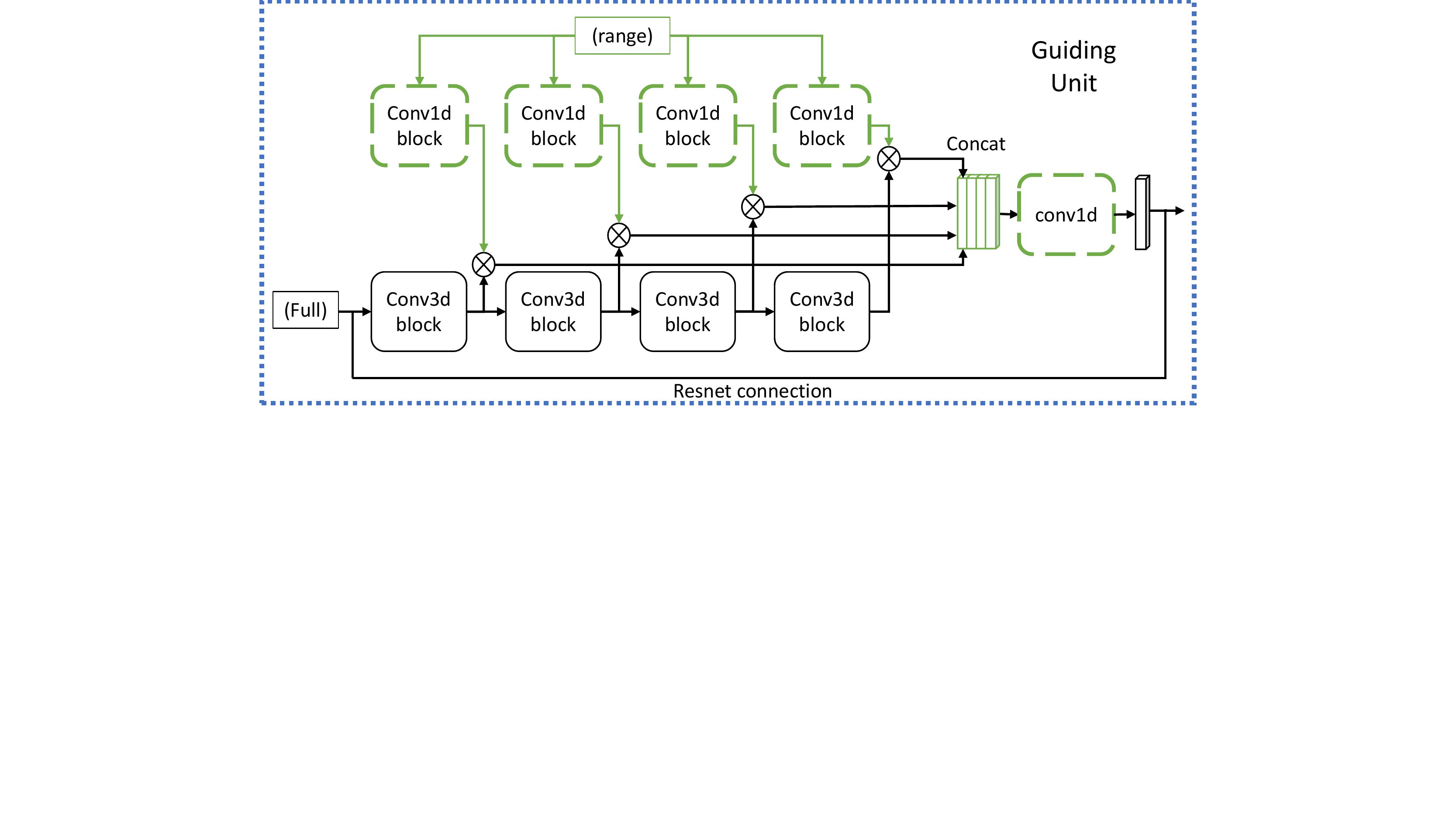}
\end{center}
 \vspace{-4.8cm}
  \caption{{\bf Illustration of the Guiding Unit:} On the range input we use four separate 1D convolutions. The output of each one is multiplied by the output of its parallel convolution layer on the full input and then concatenated together. In order to achieve the original number of features, we apply 1D convolution with a residual \cite{resnet} connection.}
\label{fig:weighting_unit}
\end{figure*}

% Within the weighting unit, we use four separate 1D convolutions, all directly from the range input. The output of each one is multiply by the output of its parallel convolution layer on the full input and then concatenate all four together. In order to achieve the original number of features, we then use 1D convolution with a ResNet\cite{resnet} connection.  

%% file: 03_method.tex
% \input{Images/cylin/range_center}
%-------------------------------------------------------------------------------------------
\input{Images/cylin/range_center}
\section{Method}
% As presented in the above section, in order to take advantage of the many benefits of the Cylinder voxelization, we need to surmount the challenges raised.

% The network has three main learning blocks: Range-Guided Backbone, Region Proposal Network (RPN), and Multi-group Center Head.

%The network has three learning blocks: (1) Range-Guided Backbone which receives the 3D voxels input and by using guiding units, is guided by range to output a flattened 2D map, (2) Region Proposal Network and (3) a Multi-group Center Head which output a heatmap by class and the rest of the required parameters per each detection.

Our network consists of three learning blocks: (1) Range-Guided Backbone, which receives the 3D voxelized input and constructs an output of a flattened 2D map by utilizing our novel guiding units, (2) Region Proposal Network and (3) a Multi-Group Center Head that outputs a heat-map per class and a variety of predicted parameters for each detected object. Adapting the entire solution to handle the cyclical nature of the $\theta$ axis called for modifications across all three.
%First, we address the circularity of the $\theta$ axis of the neural network by substituting zero-padding with circular padding for the $\theta$-axis in all of the convolution layers throughout the entire network, both 2D and 3D.
% which receives the 3D voxels as the input and output a 2D map to the next one which is the Region Proposal Network (RPN), and the third and last is the center head. We will  

%-----------------------------------------------------------------
\paragraph{Input:}
%Although the natural output of a LiDAR sensor is given in range and two angles, the nuScenes dataset provides the point cloud sweeps in the Cartesian coordinate system. Additionally, each point has two other features: intensity, which is taken directly from the sensor, and $\Delta t$, which relates to the lag-time between each sweep to its key-frame sweep. We followed the rules of the nuScenes detection benchmark by taking ten sequential samples (totaling 0.5 seconds). We then quantize it according to cylindrical voxel resolution and average on each voxel along all features.   

Although the raw output of a LiDAR sensor is given in range and two angles, the nuScenes dataset provides the point cloud sweeps already transformed to Cartesian coordinates. Each point is assigned with two additional features: intensity, which is taken directly from the sensor, and $\Delta t$, which relates to the lag-time between each sweep to its key-frame sweep. We followed the rules of the nuScenes detection benchmark by taking ten sequential samples (totaling 0.5 seconds). We then quantize it according to the cylindrical voxel resolution and average on each voxel across all features.

%-----------------------------------------------------------------
\paragraph{Range-Guided Backbone:}  
In the first block of our proposed solution, we introduce the range-guided backbone. Since similar objects appear across different bins depending on the range, we propose to control the convolutions along the main pipeline using a range guided mechanism.

%Being guided by range when handling a point cloud data in the outdoor scene is true regardless of which coordinate system is used, but also hold a key factor when handling a map with various size voxels.
% As we have previously put forth, we designed a novel range guided backbone, which can easily be added to other pipelines. Before delving into specific technical details, let's discuss the two primary reason why one should change its network backbone to be range-guided.
The motivation for range guidance is two-fold; 
First, the LiDAR transmitter releases rapid pulses of laser light across a pre-defined spatial pattern. Using a synced sensor, it measures the time it takes for each pulse to reflect back from the object it hits. Then, we receive multiple hits per object, which depends on its distance. Meaning, nearby objects will be over-represented in contrast to far away objects, as can be seen in Figure \ref{fig:avg_points} leading to significant deviation in close and distant objects statistics.
The second reason relates to the voxel size. As noted earlier (see Section \ref{challenges}), the voxel size in BEV increases by the range in Cylindrical coordinates. Which means nearby objects need a wider receptive field than far away objects.

The main \emph{guided} backbone was inspired by CBGS \cite{cbgs} who followed the work of SECOND\cite{second}. 
The range \emph{guiding} backbone is made of a convolution pipeline sourced by range only matrix (meaning only the range feature of each voxel), controlling the number and weights of consecutive convolutions running on the main pipeline. See Figure \ref{fig:only_backbone} for visual aid.

%The new backbone needs to achieve two objectives. The first is to allow different reception fields, the second is to be guided by range. In order to do so, we first added another input of solely the range feature. 
Specifically, the range backbone is parallel to the main pipeline. In each guiding unit, the output of each 3D convolution running on the main pipeline is multiplied by the output of a 1 x 1 convolution which directly operates on the range matrix. The product of four such multiplications are concatenated and then reduced in dimensions by a 1 x 1 convolution operator.
The design of the guiding unit allows the neural network to learn different receptive fields, while the 3D convolution layers in between contribute to its learning capabilities of the spatial information of neighboring voxels. In our Cylindrical network, this backbone will serve mainly the far range where the receptive field needs to be smaller.

% The range feature branch is parallel to the main original branch and we apply both 3D convolutions and 1D convolutions on it. 3D convolutions are applied with stride in order to obtain spatial information from the neighboring voxels, while 1D convolutions are applied inside the weighting block where the map size remains fixed. 

%-------------------------------------------------------------------------------------------
\paragraph{RPN:}
%We found the RPN used by CBGS who followed VoxelNet\cite{cbgs,voxelnet} is suitable for our network.
We have found that the RPN adopted by CBGS after following the VoxelNet \cite{cbgs,voxelnet} method is suitable for our network as well.
% The RPN is using two levels of features: on the first level, we use multiple 2D convolutions with stride 1 and on the second level, we apply a single 2D convolution with stride 2 and multiple 2D convolutions with stride 1. Finally, a single deconvolution is applied on the second level to return it to the original map size. We then concatenate them together to obtain a higher feature map.

The RPN consists of two levels of features. On both levels, we use multiple 2D convolutions, but on one of them, we first use a single convolution with stride 2, and in the end, we apply deconvolution to return it to the original map size. Both levels then get concatenated together to obtain a higher feature map.

%We follow VoxelNet and CBGS\cite{voxelnet,cbgs} and adopt a regular 2D convolution followed by deconvolution in order to obtain a higher feature map by using aggregation. 
%----------------------------------------------------------------------------------------------
%-------------------------------------------------------------------------------------------

\paragraph{Multi Group Center Head:} \label{mgch}
%We follow CenterPoint\cite{centerpoint} head which uses a K-channel heatmap to indicate the centers of the K-class objects. In addition to the heatmap, the network output ten different scalars for each detection: $(dx,dy)$ which is the delta to the object's center from the heatmap coordinates, $(w,l,h)$ which indicate the object size, $(Z)$ for the object's center in the Z-axis, $(v_x,v_y)$ for the object's velocity in X and Y axes respectively and $(cos\theta_{dir}, sin\theta_{dir})$ where $\theta_{dir}$ is the heading angle of the object. 
%In order to overcome the orientation adaption challenge we performed modification.
%In order to address the self-centered issue, we preformed the following modifications:

For this block, we follow the head architecture of CenterPoint \cite{centerpoint}. In order to make use of the block, some adjustments were needed to overcome the aforementioned challenges using Cylindrical coordinates.

The original head outputs a K-channel 2D heatmap to indicate the centers of the K-class objects in the x-y plane. In addition to the heatmap, the network outputs ten different scalars for each detection: $(dx,dy)$ which is the delta to the object's center from the heatmap coordinates, $(w,l,h)$ which indicate the object width, length and height, $(z_{center})$ for the object's center in the Z-axis, $(v_x,v_y)$ for the object's velocity in X and Y axes respectively and $(\cos\theta_{dir}, \sin\theta_{dir})$ where $\theta_{dir}$ is the heading angle of the object.

%----------------------------------------------------------------------------------------------
The ground truth heatmaps for training are built by rendering a Gaussian around each object's center. CenterPoint \cite{centerpoint} uses symmetric 2D Gaussian since the X and Y axes share the same resolution, and calculate the radius using the following equation:
\begin{equation} \label{eq:sigma}
\sigma=\max(f(wl,r),\tau)
\end{equation}
Where $f$ is the radius function defined in CornerNet \cite{cornernet}, $\tau$ is the smallest allowable Gaussian radius and $w,l$ are the width and length of the detected object in \emph{number} of voxels, meaning the result of $w$ and $l$ divided by the voxel's size.

In our setup, we can not use symmetric Gaussian anymore thus we split it into 2D Gaussian with different radii for each axis which are calculated separately by equation (\ref{eq:sigma}).
We then calculate the size of a voxel in the theta-axis by the following formula:
\begin{equation}
    V_{\theta size} = 2r_{center}\sin(\theta_{center}/2)
\end{equation}
Where $r_{center}$ , and $\theta_{center}$ are the coordinates of the object's center in the r and $\theta$ dimensions respectively.

\paragraph{} While the transformation of the heatmap coordinates and the deltas from Cartesian to Cylindrical coordinates is quite simple, the heading angle and velocity outputs are much more complex and require a deep understanding of the network's view.

%\begin{equation}
%    r=\sqrt{x^2+y^2}, \theta=\arctan(x/y)
%\end{equation}
% Where we use x as the numerator and y as the denominator in order to be aligned with the nuScenes notation.
The $\theta_{dir}$ in which our Cylinder-network observes the object depends not only on the $\theta_{dir}$ of the object but also on the $\theta_{center}$ of the object. An object facing forward with $\theta_{center}=\pi/2$ will be interpreted on the 2d map the same way an object-centered right in front of the ego-vehicle and facing $\theta_{dir}=-\pi/2$ will see Figure \ref{fig:dif_range}. Therefore, the $\theta_{dir}$ we wish to learn is as follows: 
\begin{equation} \label{equation:dir}
    \Bar \theta_{dir} = \theta_{dir} - \theta_{center}
\end{equation}
Where $\theta_{center}$ is the $\theta$ coordinate of the object's center.

From the calculation of $\Bar \theta_{dir}$, we gain significant added value over the Cartesian coordinate system. As demonstrated in Figure \ref{fig:points_on_car}, points are distributed differently between similar cars with respect to their $\theta_{center}$. Although the two cars share identical $\theta_{dir}$ in the classic system, the points are scattered on different parts of each vehicle. As the blue bound car is mostly "hit" on its front, the points "hitting" the green marked car mainly concentrate on its left side. In our modified system, for instance, when most of the object's cloud points cover its rear, the value of its $\Bar \theta_{dir}=0$, when the majority of the points appear on its right side, $\Bar \theta_{dir}=\pi/2$, and so forth. We believe it allows our system a more adequate comprehension of the scene.
\input{Images/points/car_point}

Now we can separate the new $\Bar \theta_{dir}$ into two orthogonal components:
\begin{equation} \label{equation:dir_learn}
\begin{split}
    \theta_{\theta} = \sin(\Bar \theta_{dir}) \\
    \theta_{r} = \cos(\Bar \theta_{dir})
\end{split}
\end{equation}
For each detection, one is also required to output the object's velocity. As opposed to $\theta$, the required velocity for this benchmark is already divided into the classical coordinates, meaning the metrics of this benchmark evaluates $v_x$ and $v_y$ and reported as ground truth. Therefore an extra step is required:  
\begin{equation}
\begin{split}
    V_{abs} &= \sqrt{v_{x}^2 + v_{y}^2} \\
         V_{dir} &= \arctan\left( \frac{v_{x}}{v_{y}}
         \right)
\end{split}
\end{equation}

Where we use $V_x$ as the numerator and $V_y$ as the denominator to be aligned with nuScenes coordinate system.
\newline{}We then apply the same transformation we did in equation (\ref{equation:dir}) for velocity:
\begin{equation} \label{equation:velo}
    \Bar \theta_{velocity} = V_{dir} - \theta_{center}
\end{equation}

We then to divide that into two learnable outputs according to our self-cylindrical coordinates as before:
\begin{equation} \label{equation:velo_learn}
\begin{split}
    V_{\theta} &= V_{abs}\sin(\Bar \theta_{velocity})\\
    V_{r} &= V_{abs}\cos(\Bar \theta_{velocity}).
\end{split}
\end{equation}

%--------------------------------------------------------------------------------------------

\input{Images/tables/val_results}
\input{Images/tables/test_result2}
%--------------------------------------------------------------------------------------------
\paragraph{Losses:} Following CenterPoint\cite{centerpoint} we use $L_1$ regression for the local offset $(dr,d\theta$), orientation $(\theta_{r}, \theta_{\theta})$ as declared on equation (\ref{equation:dir_learn}) and velocity $(V_{r}, V_{\theta})$ as declared on equation (\ref{equation:velo_learn}). Additionally we use log-space $L_1$ regression for size $(w,l,h)$ and $z_{center}$. To supervise the heatmap we use a focal loss\cite{focal} with $\alpha=2$ and $\beta=4$.

The overall loss of our network is as follows:
\begin{equation}
    L=L_{heatmap}+\lambda_{reg}L_{reg}
\end{equation}
Where $L_{heatmap}$ is the focal loss for the heatmap and $L_{reg}$ and $\lambda_{reg}$ are the regression losses mentioned above and their loss weight accordingly.
%-----------------------------------------------------------------------------------------
\input{Images/tables/ablation_all}

\paragraph{Circular Padding:} We address the circularity of the $\theta$ axis of the neural network by substituting zero-padding with circular padding for the $\theta$-axis in all of the convolution layers throughout the entire network, both 2D and 3D.

\paragraph{} By applying all of the above methodologies, we have completely converted the network to operate on self-cylindrical coordinates.

%\[
%L_{hm} = -\frac{1}{N} 
%  \begin{cases} 
%   (1-\hat{Y}_{p,k})^\alpha\log(\hat{Y}_{p,k}) & \text{if } {Y}_{p,k}=1 \\
%   (1-Y_{p,k})^\beta(\hat{Y}_{p,k})^\alpha\log(1-\hat{Y}_{p,k})      & o.w
%  \end{cases}
%\]

%% file: Images/cylin/range_center.tex
\begin{figure*}[hbt!]
\begin{center}
\textbf{Cylinder Map to Input Array for Objects with same \bm{$\Bar \theta_{dir}$}}
\includegraphics[width=0.9\linewidth]{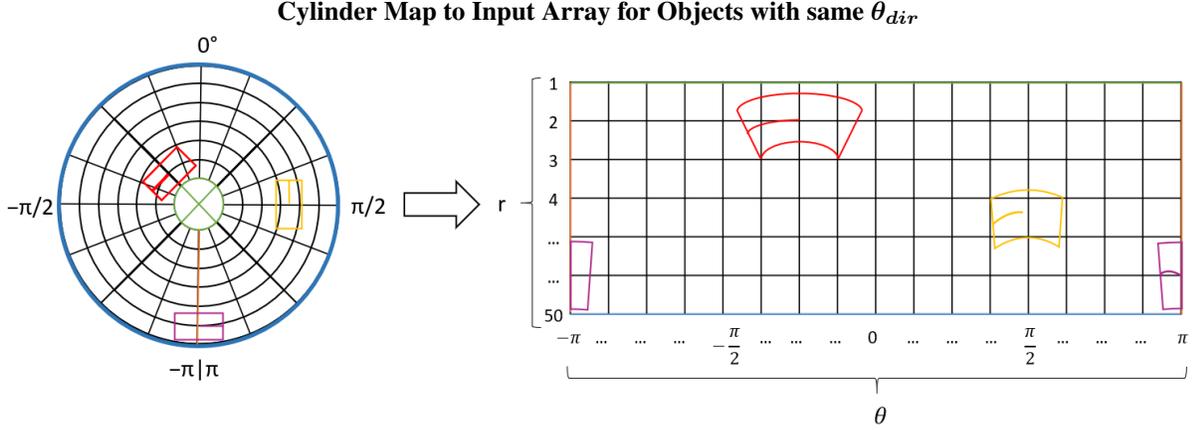}
\caption{Illustration of the Cylindrical map translated into the network's 2D array. Here shown three identical cars of the same size, but with different centers. They also have different $\theta_{dir}$ but the same $\Bar \theta_{dir}$ (defined in equation (\ref{equation:dir})), as viewed from the sensor's perspective. We can also observe here that the closer the vehicle is to the sensor, the more voxels in the $\theta$ axis it occupies. We also see here the circularity character of the $\theta$-axis where the magenta car is spread on both sides of the 2D array.}
\label{fig:dif_range}
\end{center}
\end{figure*}

%% file: Images/points/car_point.tex
\begin{figure}[htb!]
\begin{center}
    \textbf{}
    \includegraphics[width=8cm, height=7.9cm]{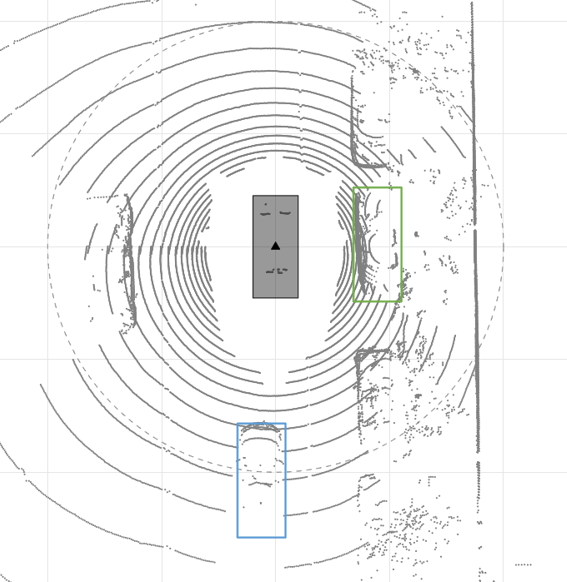}
    \caption{\textbf{A single frame of nuScenes scene, demonstrating the Cartesian system's drawback:} Although the green and blue cars share many characteristics, in particular that they both have a similar orientation - the pattern in which they are covered with point clouds is significantly different. By applying the modifications described in equation \ref{equation:dir}, this difference no longer poses a challenge to our Cylindrical network.}
        \label{fig:points_on_car}
    \end{center}
    
\end{figure}

%% file: Images/tables/val_results.tex
\begin{table*}[!t]
\centering
% \begin{center}

\begin{tabular}{|l c c c c c c ||c||} % \hspace*{} for Left-right shift of the entire table
\toprule

\textbf{Method} & \textit{mAP} & \textit{mATE} & \textit{mASE} & \textit{mAOE} & \textit{mAVE} & \textit{mAAE} & \textit{NDS} \\

\midrule            
CBGS \cite{cbgs} 
& 0.499 & 0.335
& 0.256 & 0.323
& \textbf{0.251} & 0.197
& 0.613 \\
CenterPoint \cite{centerpoint}
& \textbf{0.591} & \textbf{0.277}
& \textbf{0.251} & \textbf{0.269}
& \underline{0.258} & \underline{0.189}
& \textbf{0.671}\\
Ours
& \underline{0.576} &	\underline{0.283} &	\underline{0.253} &	\underline{0.291} &	0.268 &	\textbf{0.180} &	\underline{0.661}\\
\midrule            

%& 0.5705 & 0.2830 
%& 0.2520 & 0.2917
%& 0.2690 & 0.1726
%& 0.6584 \\
\end{tabular}
% \end{center}
\vspace{0.08cm}

\caption{ {\bf nuScenes Experiment Results on Validation set:} The results of CBGS and CenterPoint as shown on their github. Top results are in bold, second place is underlined.}
%\vspace{0.08cm}		

\vspace{-0.1cm}
\label{tables: val results}
\end{table*}	

%% file: Images/tables/test_result2.tex
\begin{table*}[!t]
\vspace{0.18cm}
\centering
% \begin{center}
\begin{tabular}{|l c c c c c c ||c|| } % \hspace*{} for Left-right shift of the entire table
\toprule

\textbf{Method} & \textit{mAP} & \textit{mATE} & \textit{mASE} & \textit{mAOE} & \textit{mAVE} & \textit{mAAE} & \textbf{NDS} \\

\midrule            
CBGS \cite{cbgs} 
& 0.528
& 0.300
& 0.247
& \underline{0.379}
& \underline{0.245}
& 0.140
& 0.633 \\
CVCNet single \cite{cvcnet}
& 0.558
& 0.300
& 0.248
& 0.431
& 0.269
& \textbf{0.119}
& 0.642 \\
CVCNet single V2 \cite{cvcnet}
& 0.553
& 0.300
& 0.244
& 0.389
& 0.268
& \underline{0.122}
& 0.644 \\
MMDetection3D \cite{mmdetection}
& 0.575
& 0.316
& 0.256
& 0.409
& \textbf{0.236}
& 0.124
& 0.653 \\
HotSpotNet-0.1m \cite{hotspot}
& \underline{0.593}
& 0.274
& \textbf{0.239}
& 0.384
& 0.333
& 0.133
& 0.660 \\
CenterPoint single\cite{centerpoint}
& \textbf{0.603}
& \textbf{0.262}
& \textbf{0.239}
& \textbf{0.361}
& 0.288
& 0.136
& \textbf{0.673} \\
\midrule  
Ours
& 0.585
& \underline{0.272}
& \underline{0.243}
& 0.383
& 0.293
& 0.126
& \underline{0.661} \\
\midrule            

%& 0.5705 & 0.2830 
%& 0.2520 & 0.2917
%& 0.2690 & 0.1726
%& 0.6584 \\
\end{tabular}
% \end{center}
\vspace{0.08cm}		
\caption{ {\bf nuScenes Experiment Results on Test set:} Top results for \textit{single model} on the LiDAR track, taken from the nuScenes website. Top results are in bold, second place is underlined} 

%\vspace{0.1cm}
\label{tables:test results}
\end{table*}	

%% file: Images/tables/ablation_all.tex
\begin{table*}[!t]
\centering
% \begin{center}

\begin{tabular}{|c c c c c c c c c c c ||c|| } % \hspace*{} for Left-right shift of the entire table
\toprule

\textbf{CC.mod.} & \textit{0.5m} & \textit{1m} & \textit{2m} & \textit{4m} & \textit{mAP}&\textit{mATE} & \textit{mASE} & \textit{mAOE} & \textit{mAVE} & \textit{mAAE} & \textit{NDS} \\

\midrule            
X &
0.468 &	0.561 &	0.618&	0.648&	0.574&	\textbf{0.282}&	\textbf{0.250}&	 0.314&	0.312&	0.185	& 0.652 \\
V &
\textbf{0.469} & \textbf{0.562} &	\textbf{0.623} &	\textbf{0.651} &	\textbf{0.576} &	0.283 &	0.253 &	\textbf{0.291} &	\textbf{0.268} &	\textbf{0.180} &	\textbf{0.661}
\\

\midrule            

%& 0.5705 & 0.2830 
%& 0.2520 & 0.2917
%& 0.2690 & 0.1726
%& 0.6584 \\
\end{tabular}
% \end{center}
\vspace{0.08cm}

\caption{ {\bf  Ablation Experiment on nuScenes Validation set:} Cylindrical-coordinate modifications (CC.mod.) refers to modification explained on subsection Multi Group Center Head under section Method~\ref{mgch} for both orientation ($\Bar \theta_{dir}$) and velocity ($\Bar \theta_{velocity}$). Notice, how our CC.mod. significantly improves performance for both, orientation (mAOE) and velocity (mAVE).}
%\vspace{0.08cm}		

\vspace{-0.1cm}
\label{tables: ablation cc all}
\end{table*}

%% file: 04_experiments.tex
\section{Experiments}
\paragraph{Dataset:}
%We evaluate our method on the large scale nuScenes dataset for 3D object detection \cite{nuscenes}. The dataset contains 700 training samples, 150 validation samples and another 150 for test, and according to the latest benchmark we are required to detect 10 different classes: car, bus, construction vehicle, trailer, truck, pedestrian, motorcycle, bicycle, traffic cone and barrier.

We evaluate our method on the nuScenes dataset for 3D object detection \cite{nuscenes}. The dataset consists of 700 training scenes, 150 validation scenes and a test set of 150 scenes, each 20s long with LiDAR frequency 20 FPS. The latest nuScenes benchmark requires the detection of 10 different classes: car, bus, construction vehicle, trailer, truck, pedestrian, motorcycle, bicycle, traffic cone and barrier.

%----------------------------------------------------------------------------
\paragraph{Evaluation Metrics:}
The nuScenes Detection Score (NDS) is a weighted sum of mean Average Precision (mAP)~\cite{pascal} 
and several True Positive (TP) metrics. The mAP is calculated as an average of four mAP with distance threshold of: (0.5m, 1m, 2m and 4m), and the TP metrics are calculated between the prediction and its \textit{matched} ground truth the  under the \textit{2m} threshold:
\begin{enumerate}
   \item Average Translation Error (\textbf{ATE}): The Euclidean distance in meters between the two centers.
   \item Average Scale Error (\textbf{ASE}): Calculated as (1 - IOU) after aligning the centers and orientation of both.
   \item Average Orientation Error (\textbf{AOE}): Smallest yaw angle difference ($\theta_{dir}$).
   \item Average Velocity Error (\textbf{AVE}): Absolute velocity error for $v_x$ and $v_y$ separately, measured in meters per second.
   \item Average Attribute Error (\textbf{AAE}): Calculated as 1 - acc, where acc is the attribute classification accuracy.
 \end{enumerate}
 
 The final NDS is calculated as follows:
 \[NDS = 5mAP+ \sum_{i=1}^{5} \max(1-TP_i, 0.0)\]

%----------------------------------------------------------------------------
\subsection{Implementation details}\label{sec:impementation_details}

\paragraph{Preprocess:}
Our implementation is based on the open-sourced code of CenterPoint\footnotemark\space implemented using the PyTorch \cite{pytorch} framework. Following the nuScenes Benchmark, we set the detection range to $[1m, 53.8m]$ for the range axis, $[-\pi, \pi]$ for the $\theta$ axis and $[-3m, 5m]$ for the Z-axis. Utilizing the structure of the cylindrical coordinates, we do not need extra spaces in our input map, in order to detect an object within the required 50m range.
\footnotetext{\url{https://github.com/tianweiy/CenterPoint}
}

With the choice of $[0.075m, \pi/600 rads, 0.2m]$ for the voxels size in each axis respectively, our network input size is 704 x 1200 x 40 voxels. Max point per voxel was set to 10 and the max number of voxels was set to 150k.
In our range-guided backbone we use four times stride 2 for the range axis, and three times for the $\theta$ axis, resulting in 88 x 300 output size. Due to the cylindrical shape of our output map, it is almost 20\% smaller than CenterPoint's~\cite{centerpoint} output map.  

We adopt the class-balanced sampling and class-grouped heads of CBGS~\cite{cbgs} to address the class imbalance of nuScenes dataset. We also conduct CBGS' augmentations on the original point-cloud in XYZ coordinates: global rotation in the range of $[-\pi/8,\pi/8]$, global scaling with random scale factor between $[0.95m, 0.95m]$, translation within the range of 0.2m for each axis and random flip for x and y-axis. 
% \input{Images/tables/ablation_cls_velo}
% \input{Images/tables/ablation_cls_orient}

%----------------------------------------------------------------------------
\paragraph{Training:}
We train the model with a batch size of 56 for 20 epochs on 8 RTX8000 GPUs.  We have chosen adamW~\cite{adamw} optimizer with one-cycle policy~\cite{onecycle}, LR max 0.0035 with division factor 10, momentum from 0.85 to 0.95, weight decay 0.01 and chose $\lambda_{reg}$ to be 0.25. All zero-padding in $\theta$ dimension was replaced with circular-padding and the size of the kernel was increased to 5 (instead of 3) along the backbone, to obtain a larger maximum reception field in this dimension.

%----------------------------------------------------------------------------
\paragraph{Testing:}
During inference, we keep the top 500 detection proposals per sample. Consequentially, we filter out all proposals with a detection score lower than 0.1 and execute Non-maximum Suppression (NMS) with an IOU threshold of 0.1 and a max output of 83 proposals per group. We use CenterPoint~\cite{centerpoint} flip-test method but instead of using three additional flips (y-flip, x-flip, and xy-flip) to the original input, we use only one (y-flip) resulting in a reduced inference time by a factor of two.
%---------------------------------------------------------------------------
\subsection{Results}\label{sec:results}
%We present our results both on Validation samples (Table \ref{tables: val results}.) and on nuScenes test set (Table \ref{tables:test results}.) via their server with hidden annotations. Our single model on the LiDAR track outperforms all methods but one. It resulted in a higher score than both the CVCNet~\cite{cvcnet} methods, the only other method that did not use Cartesian coordinates. NuScenes' velocity evaluation is currently biased towards Cartesian systems since mAVE is measured as L2 over $v_x$ and $v_y$ separately rather than evaluating the absolute velocity and velocity angle, for a better perception of the scene. Despite that, we still manage to get a high NDS.

We present our results on both nuScenes validation and test sets. The test results were obtained from nuScenes official evaluation server, where test sample annotations remain hidden. Our single model on the LiDAR track outperforms all methods but a single one. Our network achieved a higher score than both of CVCNet~\cite{cvcnet} submissions, which is the only other method that did not use Cartesian coordinates.

%----------------------------------------------------------------------------

%---------------------------------------------------------------------------
\subsection{Ablation studies}\label{sec:abelation}
We further conduct our ablation studies on the nuScenes validation set. 

%First, we show the effectiveness of our range-guided backbone to reduce the receptive field for larger distance and smaller objects. For range $[30m, 50m]$ over all classes we get NDS of 0.478 when we omit the guidance mechanism and NDS of 0.485 with the guidance mechanism. Additionally, we observe an increase in mAP in four out of the five smallest classes (i.e barrier, traffic cone, cyclist, motorcycle and pedestrian) for the entire range, resulting in decrease from 0.574 mAP to 0.576 mAP over all classes.  
%This indicates that our proposed backbone has learned to reduce the receptive field for both far ranges using the 1~x~1 convolution layers and smaller objects using the 3~x~3 convolution layers to gain spatial information.  

%The next ablation study shows the importance of our Cylindrical-coordinate modifications detailed in Multi-Group Center Head on section \ref{mgch}. Table \ref{tables: ablation cc all}. shows the result for two identical networks both training the same way and both using circular padding. The first network was fitted with velocity and orientation modifications, while the other network did not. 
%As expected, the most significant differences between the two results lie in the velocity and orientation metrics, but we also see improvements in each of the mAP metrics (i.e. 0.5m, 1m, 2m, and 4m). 

% Although the calculations made for orientation and velocity form the network outputs include the predicted $\theta_{center}$ (as can depicted from equations (\ref{equation:dir_learn}) and ($\ref{equation:velo_learn}$)) the modified network 

First, we demonstrate the significance of our Cylindrical-coordinate modifications detailed in Multi-Group Center Head on section \ref{mgch}.
Table \ref{tables: ablation cc all} shows the result for two identical networks, trained in the same way, both utilizing circular padding. The first network has employed our velocity and orientation modifications, while the other network did not. As expected, the most significant improvement in the first network's performance can be seen in the velocity and orientation metrics. Additionally, mAP metrics indicate an increase as well (i.e. 0.5m, 1m, 2m, and 4m) and we see a significant improvement in the NDS.

Moreover, we show the effectiveness of our range-guided backbone to reduce the receptive field for larger distances and for smaller objects.
For range $[30m, 50m]$ over all classes we achieve NDS of 0.485 when the guidance mechanism is applied, and 0.478 when omitted while using the heavily optimized backbone of CBGS~\cite{cbgs}. Additionally, we observe an increase in mAP in four out of the five smallest classes (i.e barrier, traffic cone, cyclist, and motorcycle) for the entire range, resulting in an increase from 0.574 mAP to 0.576 mAP over all classes.

%% file: 05_conclusions.tex
\section{Conclusions}
In this paper, we explore the potential of the Cylindrical coordinates system for representing LiDAR point clouds in outdoor scenes. We present a novel end-to-end framework for 3D object detection in those coordinates together with a novel range-guided backbone that can be readily added to other pipelines. Our method achieves powerful results on the highly challenging nuScenes dataset and lays the cornerstone for further Cylindrical methods in this domain. 

% We proposed a novel 3D detection neural network in self-cylindrical coordinates and range-guided backbone.

%% file: ms.bbl
\begin{thebibliography}{10}\itemsep=-1pt

\bibitem{sementickitti}
J. Behley, M. Garbade, A. Milioto, J. Quenzel, S. Behnke, C. Stachniss, and J.
  Gall.
\newblock {SemanticKITTI: A Dataset for Semantic Scene Understanding of LiDAR
  Sequences}.
\newblock In {\em Proc. of the IEEE/CVF International Conf.~on Computer Vision
  (ICCV)}, 2019.

\bibitem{range_google}
Alex Bewley, Pei Sun, Thomas Mensink, Dragomir Anguelov, and Cristian
  Sminchisescu.
\newblock Range conditioned dilated convolutions for scale invariant 3d object
  detection, 2020.

\bibitem{nuscenes}
Holger Caesar, Varun Bankiti, Alex~H. Lang, Sourabh Vora, Venice~Erin Liong,
  Qiang Xu, Anush Krishnan, Yu Pan, Giancarlo Baldan, and Oscar Beijbom.
\newblock nuscenes: A multimodal dataset for autonomous driving.
\newblock {\em arXiv preprint arXiv:1903.11027}, 2019.

\bibitem{pointnet}
R.~Q. {Charles}, H. {Su}, M. {Kaichun}, and L.~J. {Guibas}.
\newblock Pointnet: Deep learning on point sets for 3d classification and
  segmentation.
\newblock In {\em 2017 IEEE Conference on Computer Vision and Pattern
  Recognition (CVPR)}, pages 77--85, 2017.

\bibitem{dilated1}
L. {Chen}, G. {Papandreou}, I. {Kokkinos}, K. {Murphy}, and A.~L. {Yuille}.
\newblock Deeplab: Semantic image segmentation with deep convolutional nets,
  atrous convolution, and fully connected crfs.
\newblock {\em IEEE Transactions on Pattern Analysis and Machine Intelligence},
  40(4):834--848, 2018.

\bibitem{hotspot}
Qi Chen, Lin Sun, Zhixin Wang, Kui Jia, and Alan Yuille.
\newblock Object as hotspots: An anchor-free 3d object detection approach via
  firing of hotspots, 2020.

\bibitem{mv3d}
Xiaozhi Chen, Huimin Ma, Ji Wan, Bo Li, and Tian Xia.
\newblock Multi-view 3d object detection network for autonomous driving.
\newblock In {\em Proceedings of the IEEE Conference on Computer Vision and
  Pattern Recognition (CVPR)}, July 2017.

\bibitem{pointrcnn}
Yilun Chen, Shu Liu, Xiaoyong Shen, and Jiaya Jia.
\newblock Fast point r-cnn.
\newblock In {\em Proceedings of the IEEE/CVF International Conference on
  Computer Vision (ICCV)}, October 2019.

\bibitem{pascal}
Mark Everingham, Luc Gool, Christopher~K. Williams, John Winn, and Andrew
  Zisserman.
\newblock The pascal visual object classes (voc) challenge.
\newblock {\em Int. J. Comput. Vision}, 88(2):303–338, June 2010.

\bibitem{kitti}
Andreas Geiger, Philip Lenz, and Raquel Urtasun.
\newblock Are we ready for autonomous driving? the kitti vision benchmark
  suite.
\newblock In {\em Conference on Computer Vision and Pattern Recognition
  (CVPR)}, 2012.

\bibitem{onecycle}
Sylvain Gugger.
\newblock The 1cycle policy.
\newblock https://sgugger.github.io/the-1cycle-policy.html.

\bibitem{resnet}
Kaiming He, X. Zhang, Shaoqing Ren, and Jian Sun.
\newblock Deep residual learning for image recognition.
\newblock {\em 2016 IEEE Conference on Computer Vision and Pattern Recognition
  (CVPR)}, pages 770--778, 2016.

\bibitem{avod}
Jason Ku, Melissa Mozifian, Jungwook Lee, Ali Harakeh, and Lake~Steven
  Waslander.
\newblock Joint 3d proposal generation and object detection from view
  aggregation.
\newblock {\em Intelligent Robots and Systems (IROS)}, 2018.

\bibitem{pointpillars}
Alex~H. Lang, Sourabh Vora, Holger Caesar, Lubing Zhou, Jiong Yang, and Oscar
  Beijbom.
\newblock Pointpillars: Fast encoders for object detection from point clouds.
\newblock In {\em CVPR}, 2019.

\bibitem{cornernet}
Hei Law and Jia Deng.
\newblock Cornernet: Detecting objects as paired keypoints.
\newblock In {\em Proceedings of the European Conference on Computer Vision
  (ECCV)}, September 2018.

\bibitem{rangercnn}
Zhidong Liang, Ming Zhang, Zehan Zhang, Xian Zhao, and Shiliang Pu.
\newblock Rangercnn: Towards fast and accurate 3d object detection with range
  image representation, 2020.

\bibitem{focal}
T. {Lin}, P. {Goyal}, R. {Girshick}, K. {He}, and P. {Dollár}.
\newblock Focal loss for dense object detection.
\newblock In {\em 2017 IEEE International Conference on Computer Vision
  (ICCV)}, pages 2999--3007, 2017.

\bibitem{adamw}
Ilya Loshchilov and Frank Hutter.
\newblock Decoupled weight decay regularization.
\newblock In {\em 7th International Conference on Learning Representations,
  {ICLR} 2019}, 2019.

\bibitem{lasernet}
Gregory~P. Meyer, Ankit Laddha, Eric Kee, Carlos Vallespi-Gonzalez, and Carl~K.
  Wellington.
\newblock Lasernet: An efficient probabilistic 3d object detector for
  autonomous driving.
\newblock In {\em The IEEE Conference on Computer Vision and Pattern
  Recognition (CVPR)}, 2019.

\bibitem{pytorch}
Adam Paszke, Sam Gross, Francisco Massa, Adam Lerer, James Bradbury, Gregory
  Chanan, Trevor Killeen, Zeming Lin, Natalia Gimelshein, Luca Antiga, Alban
  Desmaison, Andreas Kopf, Edward Yang, Zachary DeVito, Martin Raison, Alykhan
  Tejani, Sasank Chilamkurthy, Benoit Steiner, Lu Fang, Junjie Bai, and Soumith
  Chintala.
\newblock Pytorch: An imperative style, high-performance deep learning library.
\newblock In H. Wallach, H. Larochelle, A. Beygelzimer, F. d\textquotesingle
  Alch\'{e}-Buc, E. Fox, and R. Garnett, editors, {\em Advances in Neural
  Information Processing Systems 32}, pages 8024--8035. Curran Associates,
  Inc., 2019.

\bibitem{pointnet++}
Charles~Ruizhongtai Qi, Li Yi, Hao Su, and Leonidas~J Guibas.
\newblock Pointnet++: Deep hierarchical feature learning on point sets in a
  metric space.
\newblock In I. Guyon, U.~V. Luxburg, S. Bengio, H. Wallach, R. Fergus, S.
  Vishwanathan, and R. Garnett, editors, {\em Advances in Neural Information
  Processing Systems}, volume~30, pages 5099--5108. Curran Associates, Inc.,
  2017.

\bibitem{cvcnet}
Ernest~Cheung Qi~Chen, Lin~Sun and Alan~L. Yuille.
\newblock Every view counts: Cross-view consistency in 3d object detection with
  hybrid-cylindrical-spherical voxelization.
\newblock In {\em Advances in Neural Information Processing Systems 33
  pre-proceedings (NeurIPS)}, 2020.

\bibitem{pvrcnn}
Shaoshuai Shi, Chaoxu Guo, Li Jiang, Zhe Wang, Jianping Shi, Xiaogang Wang, and
  Hongsheng Li.
\newblock Pv-rcnn: Point-voxel feature set abstraction for 3d object detection.
\newblock In {\em IEEE/CVF Conference on Computer Vision and Pattern
  Recognition (CVPR)}, June 2020.

\bibitem{partA2}
Shaoshuai Shi, Zhe Wang, Jianping Shi, Xiaogang Wang, and Hongsheng Li.
\newblock From points to parts: 3d object detection from point cloud with
  part-aware and part-aggregation network.
\newblock {\em IEEE Transactions on Pattern Analysis and Machine Intelligence},
  PP:1--1, 02 2020.

\bibitem{dilated2}
P. {Wang}, P. {Chen}, Y. {Yuan}, D. {Liu}, Z. {Huang}, X. {Hou}, and G.
  {Cottrell}.
\newblock Understanding convolution for semantic segmentation.
\newblock In {\em 2018 IEEE Winter Conference on Applications of Computer
  Vision (WACV)}, pages 1451--1460, 2018.

\bibitem{second}
Yan Yan, Yuxing Mao, and Bo Li.
\newblock Second: Sparsely embedded convolutional detection.
\newblock {\em Sensors}, 18(10):3337, Oct 2018.

\bibitem{pixor}
Bin Yang, Wenjie Luo, and Raquel Urtasun.
\newblock Pixor: Real-time 3d object detection from point clouds.
\newblock In {\em Proceedings of the IEEE Conference on Computer Vision and
  Pattern Recognition (CVPR)}, June 2018.

\bibitem{centerpoint}
Tianwei Yin, Xingyi Zhou, and Philipp Krähenbühl.
\newblock Center-based 3d object detection and tracking, 2020.

\bibitem{mmdetection}
Wenwei Zhang, Kai Chen, Zhe Wang, Jianping Shi, and Chen~Change Loy.
\newblock Mmdetection3d.
\newblock https://github.com/open-mmlab/mmdetection3d.

\bibitem{polarnet}
Yang Zhang, Zixiang Zhou, Philip David, Xiangyu Yue, Zerong Xi, Boqing Gong,
  and Hassan Foroosh.
\newblock Polarnet: An improved grid representation for online lidar point
  clouds semantic segmentation.
\newblock In {\em Proceedings of the IEEE/CVF Conference on Computer Vision and
  Pattern Recognition (CVPR)}, June 2020.

\bibitem{voxelnet}
Y. {Zhou} and O. {Tuzel}.
\newblock Voxelnet: End-to-end learning for point cloud based 3d object
  detection.
\newblock In {\em 2018 IEEE/CVF Conference on Computer Vision and Pattern
  Recognition}, pages 4490--4499, 2018.

\bibitem{cbgs}
Benjin Zhu, Zhengkai Jiang, Xiangxin Zhou, Zeming Li, and Gang Yu.
\newblock Class-balanced grouping and sampling for point cloud 3d object
  detection, 2019.

\end{thebibliography}
